# Value-Directed Belief State Approximation for POMDPs

**Pascal Poupart**
Department of Computer Science
University of British Columbia
Vancouver, BC V6T 1Z4
*ppoupart@cs.toronto.edu*

**Craig Boutilier**
Department of Computer Science
University of Toronto
Toronto, ON M5S 3H5
*cebly@cs.toronto.edu*

## Abstract

We consider the problem belief-state monitoring for the purposes of implementing a policy for a partially-observable Markov decision process (POMDP), specifically how one might approximate the belief state. Other schemes for belief-state approximation (e.g., based on minimizing a measure such as KL-divergence between the true and estimated state) are not necessarily appropriate for POMDPs. Instead we propose a framework for analyzing *value-directed* approximation schemes, where approximation quality is determined by the expected error in utility rather than by the error in the belief state itself. We propose heuristic methods for finding good *projection schemes* for belief state estimation—exhibiting anytime characteristics—given a POMDP value function. We also describe several algorithms for constructing bounds on the error in decision quality (expected utility) associated with acting in accordance with a given belief state approximation.

## 1 Introduction

Considerable attention has been devoted to partially-observable Markov decision processes (POMDPs) [15, 17] as a model for decision-theoretic planning. Their generality allows one to seamlessly model sensor and action uncertainty, uncertainty in the state of knowledge, and multiple objectives [1, 4]. Despite their attractiveness as a conceptual model, POMDPs are intractable and have found practical applicability in only limited special cases.

Much research in AI has been directed at exploiting certain types of problem structure to enable value functions for POMDPs to be computed more effectively. These primarily consist of methods that use the basic, explicit state-based representation of planning problems [5]. There has, however, been work on the use of *factored representations* that resemble classical AI representations, and algorithms for solving POMDPs that exploit this structure [2, 8]. Representations such as dynamic Bayes nets (DBNs) [7] are used to represent actions and structured representations of value functions are produced. Such models are important because they allow one to deal (potentially) with problems involving a large number of states (exponential in the number of variables) without explicitly manipulating states, instead reasoning directly with the factored representation.

Unfortunately, such representations do not automatically translate into effective policy implementation: given a POMDP value function, one must still maintain a *belief state* (or distribution over system states) *online* in order to implement the policy implicit in the value function. Belief state maintenance, in the worst case, has complexity equal to the size of the state space (exponential in the number of variables), as well. This is typically the case even when the system dynamics can be represented compactly using a DBN, as demonstrated convincingly by Boyen and Koller [3]. Because of this, Boyen and Koller develop an approximation scheme for monitoring dynamical systems (as opposed to POMDP policy implementation); intuitively, they show that one can decompose a process along lines suggested by the DBN representation and maintain bounded error in the estimated belief state. Specifically, they approximate the belief state by *projection*, breaking the joint distribution into smaller pieces by marginalization over subsets of variables, effectively discounting certain dependencies among variables.

In this paper, we consider approximate belief state monitoring for POMDPs. We assume that a POMDP has been solved and that a value function has been provided to us in a factored form (as we explain below). Our goal is to determine a projection scheme, or decomposition, so that approximating the belief state using this scheme hinders the ability to implement the optimal policy as little as possible. Our scheme will be quite different from Boyen and Koller's since our aim is *not* to keep the approximate belief state as "close" to the true belief state as possible (as measured by KL-divergence). Rather we want to ensure that decision quality is sacrificed as little as possible.

In many circumstances, this means that small correlations need to be accounted for, while large correlations can be ignored completely. As an example, one might imagine a process in which two parts are stamped from the same machine. If the machine has a certain fault, both parts have a high probability of being faulty. Yet if the decisions for subsequent processing of the parts are independent, the fact that the fault probabilities for the parts are dependent is irrelevant. We can thus project our belief state into two independent subprocesses with no loss in decision quality. *Assuming* the faults are independent causes a large "error" in the belief state; but this has no impact on subsequent decisions or even expected utility assessment. Thus we need not concern ourselves with this "error." In contrast, very small dependencies, when marginalized, may lead to very small "error" in the belief state; yet this small error can have severe consequences on decision quality.

Because of this, while Boyen and Koller's notion of projection offers a very useful tool for belief state approximation, the model and analysis they provide cannot be applied usefully to POMDPs. For example, in [14] this model is integrated with a (sampling-based) search tree approach to solving POMDPs. Because the error in decision quality is determined as a function of the worst-case decision quality with respect to actual belief state approximation error, the bounds are unlikely to be useful in practice. We strongly believe estimates of decision quality error should be based on direct information about the value function.

In this paper we provide a theoretical framework for the analysis of *value-directed belief state approximation (VDA)* in POMDPs. The framework provides a novel view of approximation and the errors it induces in decision quality. We use the value function itself to determine which correlations can be "safely" ignored when monitoring one's belief state. Our framework offers methods for bounding (reasonably tightly) the error associated with a given projection scheme. While these methods are computationally intensive—requiring in the worst case a quadratic increase in the solution time of a POMDP—we argue that this *offline* effort is worthwhile to enable fast *online* implementation of a policy with bounded loss in decision quality. We also suggest a heuristic method for choosing good projection schemes given the value function associated with a POMDP. Finally, we discuss how our techniques can also be applied to approximation methods other than projection (e.g., aggregation using density trees [13]).

## 2 POMDPs and Belief State Monitoring

### 2.1 Solving POMDPs

A partially-observable Markov decision process (POMDP) is a general model for decision making under uncertainty. Formally, we require the following components: a finite state space $\mathcal{S}$; a finite action space $\mathcal{A}$; a finite observation space $\mathcal{Z}$; a transition function $T : \mathcal{S} \times \mathcal{A} \to \Delta(\mathcal{S})$; an observation function $O : \mathcal{S} \times \mathcal{A} \to \Delta(\mathcal{Z})$; and a reward function $R : \mathcal{S} \to \mathbf{R}$.[1] Intuitively, the transition function $T(s, a)$ determines a distribution over next states when an agent takes action $a$ in state $s$—we write $\Pr(s, a, t)$ to denote the probability that state $t$ is reached. This captures uncertainty in action effects. The observation function reflects the fact that an agent cannot generally determine the true system state with certainty (e.g., due to sensor noise)—we write $\Pr(s, a, z)$ to denote the probability that observation $z$ is made at state $s$ when action $a$ is performed. Finally $R(s)$ denotes the immediate reward associated with $s$.[2]

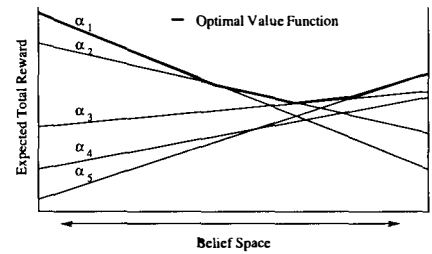

Figure 1: Geometric View of Value Function

The rewards obtained over time by an agent adopting a specific course of action can be viewed as random variables $R^{(t)}$. Our aim is to construct a *policy* that maximizes the expected sum of discounted rewards $E(\sum_{t=0}^{\infty} \gamma^t R^{(t)})$ (where $\gamma$ is a discount factor less than one). It is well-known that an optimal course of action can be determined by considering the fully-observable *belief state MDP*, where *belief states* (distributions over $\mathcal{S}$) form states, and a policy $\pi : \Delta(\mathcal{S}) \to \mathcal{A}$ maps belief states into action choices. In principle, dynamic programming algorithms for MDPs can be used to solve this problem; but a practical difficulty emerges when one considers that the belief space $\Delta(\mathcal{S})$ is an $|\mathcal{S}|-1$-dimensional continuous space. A key result of Sondik [17] showed that the value function $V$ for a finite-horizon problem is piecewise-linear and convex and can be represented as a finite collection of $\alpha$-vectors.[3] Specifically, one can generate a collection $\aleph$ of $\alpha$-vectors, each of dimension $|\mathcal{S}|$, such that $V(b) = \max_{\alpha \in \aleph} b\alpha$. Figure 1 illustrates a collection of $\alpha$-vectors with the upper surface corresponding to $V$. Furthermore, each $\alpha \in \aleph$ has a specific action associated with it; so given belief state $b$, the agent should choose the action associated with the maximizing $\alpha$-vector.

Insight into the nature of POMDP value functions, which will prove critical in the methods we consider in the next section, can be gained by examining Monahan's [15] method for solving POMDPs. Monahan's algorithm proceeds by producing a sequence of $k$-stage-to-go value functions $V^k$, each represented by a set of $\alpha$-vectors $\aleph^k$. Each $\alpha \in \aleph^k$ denotes the value (as a function of the belief state) of executing a $k$-step *conditional plan*. More precisely, let the $k$-step *observation strategies* be the set $OS^k$ of mappings $\sigma : \mathcal{Z} \to \aleph^{k-1}$. Then each $\alpha$-vector in $\aleph^k$ corresponds to the value of executing some action $a$ followed by implementing some $\sigma \in OS^k$; that is, it is the value of doing $a$, and executing the $k - 1$-step plan associated with the $\alpha$-vector $\sigma(z)$ if $z$ is observed. Using $CP(\alpha)$ to denote this plan, we have that $CP(\alpha) = \langle a; \text{if } z_i, CP(\sigma(z_i)) \forall z_i \rangle$. We informally write this as $\langle a; \sigma \rangle$. We write $\alpha(\langle a; \sigma \rangle)$ to denote the $\alpha$-vector reflecting the value of this plan.

Given $\aleph^k$, $\aleph^{k+1}$ is produced in two phases. First, the set of vectors corresponding to all action-observation policies is constructed (i.e., for each $a \in \mathcal{A}$ and $\sigma \in OS^{k+1}$, the vector $\alpha$ denoting the value of plan $\langle a, CP(\sigma(z_i)) \rangle$ is added to $\aleph^{k+1}$). Second, this set is pruned by removing all *dominated* vectors. This means that those vectors $\alpha$ such that $b \cdot \alpha$ is *not maximal* for any belief state $b$ are removed from $\aleph^{k+1}$. In Figure 1, $\alpha_4$ is dominated, playing no useful role in the representation of $V$, and can be pruned. Pruning is implemented by a series of linear programs. Refinements of this approach are possible that eliminate (or reduce) the need for pruning by directly identifying only $\alpha$-vectors that are non-dominated [17, 6, 4]. Other algorithms, such as incremental pruning [5], are similar in spirit to Monahan's approach, but cleverly avoid enumerating all observation policies. A finite $k$-stage POMDP can be solved optimally this way and a finite representation of its value function is assured. For infinite-horizon problems, a $k$-stage solution can be used to approximate the true value function (error bounds can easily be derived based on the differences between successive value functions).

---

[1]$\Delta(X)$ denotes the set of distributions over finite set $X$.

[2]Action costs are ignored to keep the presentation simple.

[3]For infinite-horizon problems, a finite collection may not be sufficient [18], but will generally offer a good approximation.

One difficulty with these classical approaches is the fact that the $\alpha$-vectors may be difficult to manipulate. A system characterized by $n$ random variables has a state space size that is exponential in $n$. Thus manipulating a single $\alpha$-vector may be intractable for complex systems.[4] Fortunately, it is often the case that an MDP or POMDP can be specified very compactly by exploiting structure (such as conditional independence among variables) in the system dynamics and reward function [1]. Representations such as dynamic Bayes nets (DBNs) [7] can be used to great effect; and schemes have been proposed whereby the $\alpha$-vectors are computed directly in a factored form by exploiting this representation.

Boutilier and Poole [2], for example, represent $\alpha$-vectors as decision trees in implementing Monahan's algorithm. Hansen and Feng [8] use algebraic decision diagrams (ADDs) as their representation in their version of incremental pruning.[5] The empirical results in [8] suggest that such methods can make reasonably large problems solvable. Furthermore, factored representations will likely facilitate good approximation schemes. There is no reason in principle that the other algorithms mentioned cannot be adapted to factored representations as well.

### 2.2 Belief State Monitoring

Even if the value function can be constructed in a compact way, the implementation of the optimal policy requires that the agent maintains its belief state over time. The *monitoring problem* itself is not generally tractable, since each belief state is a vector of size $|\mathcal{S}|$. Given a compact representation of system dynamics and sensors in the form of DBN, one might expect that monitoring may become tractable using standard belief net inference schemes. Unfortunately, this is generally not the case. Though variables may be initially independent (thus admitting a compact representation of a distribution), and though at each time step only a small number of variables become correlated, over time these correlations "bleed through" the DBN, rendering most (if not all) variables dependent after a time. Thus compact representation of belief state is typically impossible.

Boyen and Koller [3] have devised a clever approximation scheme for alleviating the computational burden of monitoring. In this work, no POMDP is used, but rather a stationary process, represented in a factored manner (e.g., using a DBN), is assumed. This might, for example, be the process induced by adopting a fixed policy. Intuitively, they consider *projection schemes* whereby the joint distribution is approximated by projecting it onto a set of subsets of variables. It is assumed that these subsets partition the variable set. For each subset, its marginal is computed; the approximate belief state is formed by assuming the subsets are independent. Thus only variables within the same subset can remain correlated in the approximate belief state. For instance, if there are 4 variables $A$, $B$, $C$ and $D$, the projection scheme $\{AB, CD\}$ will compute the marginal distributions for $AB$ and $CD$. The resulting approximate belief state, $\widehat{P}(ABCD) = P(AB)P(CD)$, has a compact, factored representation given by the distribution of each marginal.

Formally, we say a *projection scheme* $S$ is a set of subsets of the set of state variables such that each state variable is in some subset. This allows marginals with overlapping subsets of variables (e.g., $\{ABC, BCD\}$). We view strict partitioning as a special type of projection. Some schemes with overlapping subsets may not be computationally useful in practice because it may not be possible to easily generate a joint distribution from them by building a clique tree. We therefore classify as *practical* those projection schemes for which a joint distribution is easily obtained. Assuming that belief state monitoring is performed using the DBN representing the system dynamics (see [10, 12] for details on inference with DBNs), we obtain belief state $b^{t+1}$ from $b^t$ using the following steps: (a) construct a clique tree encoding the variable dependencies of the system dynamics (for a specific action and observation) and the correlations that have been preserved by the marginals representing $b^t$; (b) initialize the clique tree with the transition probabilities, the observation probabilities and the (approximate, factored) joint distribution $b^t$; (c) query the tree to obtain the distribution $b^t_+$ at the next time step; and (d) project $b^t_+$ according to some practical projection scheme $S$ to obtain the collection of marginals representing $b^{t+1} = S(b^t_+)$. The complexity of belief state updating is now exponential only in the size of the largest clique rather than the total number of variables.

Boyen and Koller show how to compute a bound on the KL-divergence of the true and approximate belief states, exploiting the contraction properties of Markov processes (under certain assumptions). But direct translation of these bounds into *decision quality error* for POMDPs generally yields weak bounds [14]. Furthermore, the suggestions made by Boyen and Koller for choosing good projection schemes are designed to minimize KL-divergence, not to minimize error in expected value for a POMDP. For this rea-son, we are interested in new methods for choosing projections that are directly influenced by considerations of value and decision quality.

---

[4] The number of $\alpha$-vectors can grow exponentially in the worst case, as well; but for many problems the number remains manageable; and approximation schemes that simply bound their number have been proposed [6].

[5] ADDs, commonly used in verification, have been applied very effectively to the solution of fully-observable MDPs [9].

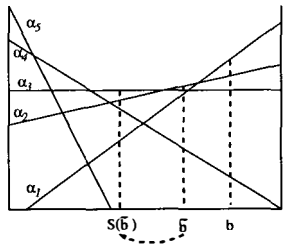

Figure 2: Relevant belief states at stage $k$

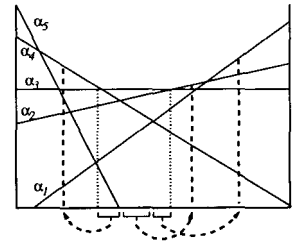

Figure 3: The Switch Set $Sw^k(\alpha_3)$ of $\alpha_3$

Other belief state approximation schemes can be used for belief state monitoring. For example, aggregation using density trees can provide a means of representing a belief state with many fewer parameters than the full joint. Our model can be applied to such schemes as well.

## 3 Error Bounds on Approximation Schemes

In this section, we assume that a POMDP has been solved and that its value function has been provided to us. We also assume that some structured technique has been used so that $\alpha$-vectors representing the value function are structured [2, 8]. We begin by assuming that we have been given an approximation scheme $S$ for belief state monitoring in a POMDP and derive error bounds associated with acting according to that approximation scheme. We focus primarily on projection, but we will mention how other types of approximation can be fit into our model. We present two techniques for bounding the error for a given approximation scheme and show that the complexity of these algorithms is similar to that of solving the POMDP, with a (multiplicative) overhead factor of $|\aleph|$.

### 3.1 Plan Switching

Implementing the policy for an infinite-horizon POMDP requires that one maintains a belief state, plugging this into the value function at each step, and executing the action associated with the maximizing $\alpha$-vector. When the belief state $b$ is approximated using an approximation scheme $S$, a suboptimal policy may be implemented since the maximizing vector for $S(b)$ will be chosen rather than the maximizing vector for $b$. Furthermore this mistaken choice of vectors (hence actions) can be compounded with each further approximation at later stages of the process. To bound such error, we first define the notion of *plan switching*. We phrase our definitions in terms of finite-horizon value functions, introducing the minor variations needed for infinite-horizon problems later.

Suppose with $k$ stages-to-go, the true belief state, had we monitored accurately to that point, is $b$. However, due to previous belief state approximations we take our current belief state to be $\tilde{b}$. Now imagine our approximation scheme has been applied at time $k$ to obtain $S(\tilde{b})$. Given $\aleph^k$, representing $V^k$, suppose the maximizing vectors associated with $b$, $\tilde{b}$ and $S(\tilde{b})$ are $\alpha_1$, $\alpha_2$ and $\alpha_3$, respectively (see Figure 2). The approximation at stage $k$ mistakenly induces the choice of the action associated with $\alpha_3$ instead of $\alpha_2$ at $\tilde{b}$; this incurs an error in decision quality of $b \cdot \alpha_2 - b \cdot \alpha_3$. While the optimal choice is in fact $\alpha_1$, the unaccounted error $b \cdot \alpha_1 - b \cdot \alpha_2$ induced by the prior approximations will be viewed as caused by the earlier approximations; our goal at this point is simply to consider the error induced by the *current* approximation.

In order to derive an error bound, we must identify, for each $\alpha \in \aleph^k$, the set of vectors $Sw^k(\alpha)$ that the agent can *switch to* by approximating its current belief state $\tilde{b}$ given that $\tilde{b}$ identifies $\alpha$ as optimal. Formally, we define

$$Sw^k(\alpha) = \{\alpha' \in \aleph^k : \exists b \forall \bar{\alpha}(b\cdot\alpha \geq b\cdot\bar{\alpha}, S(b)\cdot\alpha' \geq S(b)\cdot\bar{\alpha})\}$$

Intuitively, this is the set of vectors we could choose as maximizing (thus implementing the corresponding conditional plan) due to belief state approximation. In Figure 3, we see that $Sw^k(\alpha_3) = \{\alpha_1, \alpha_2, \alpha_4\}$. The set $Sw^k(\alpha_i)$ can be identified readily by solving a series of $O(|\aleph^k|)$ optimization problems, each testing the possibility of switching to a specific vector $\alpha_j \in \aleph^k$, formulated as the following (possibly nonlinear) program:

$$\begin{array}{ll} \max & d \\ s.t. & b \cdot (\alpha_i - \alpha_l) \geq d \quad \forall l \neq i \\ & S(b) \cdot (\alpha_j - \alpha_l) \geq d \quad \forall l \neq j \\ & \sum_s b(s) = 1 \\ & b(s) \geq 0 \quad \forall s \end{array}$$

The solution to this program has a positive objective function value whenever there is a belief state $b$ such that $\alpha_i$ is optimal at $b$, and $\alpha_j$ is optimal at $S(b)$. Note, in fact, that we need only find a positive feasible solution, not an optimal one, to identify $\alpha_j$ as an element of $Sw^k(\alpha_i)$. There are $|\aleph^k|$ switch sets to construct, so $O(|\aleph^k|^2)$ optimization problems need to be solved to determine all switch sets.

For linear approximation schemes (i.e., those in which the constraints on $S(b)$ are linear in the variables $b_i$), these problems are easily solvable linear programs (LPs). We return to linear schemes in Section 6. Unfortunately, projection schemes are nonlinear, making optimization (or identification of feasible solutions) more difficult. On the other hand, a projection scheme determines a set of linear constraints on the approximate belief state $S(b)$. For instance, consider the projection scheme $S = \{CD, DE\}$ for a POMDP with 3 binary variables. This projection imposes one linear constraint on $S(b)$ for each subset of the marginals in the projection:[6]

$$\begin{aligned} b(\emptyset) &= b'(\emptyset) & b(C) &= b'(C) \\ b(D) &= b'(D) & b(E) &= b'(E) \\ b(CD) &= b'(CD) & b(DE) &= b'(DE) \end{aligned}$$

Here $b'$ denotes $S(b)$ and $b(XY)$ denotes the cumulative probability (according to belief state $b$) of all states where $X$ and $Y$ are true. These constraints define an LP that can be used to construct a superset $\widehat{Sw}^k(\alpha_i)$ of $Sw^k(\alpha_i)$. Given scheme $S = \{M_1, \ldots, M_n\}$, we define the following LP:

$$\begin{aligned} \max \quad & d \\ \text{s.t.} \quad & b \cdot (\alpha_i - \alpha_l) \geq d & \forall l \neq i \\ & b' \cdot (\alpha_j - \alpha_l) \geq d & \forall l \neq j \\ & b'(M) = b(M) & \forall M \subseteq M_l,\ 1 \leq l \leq n \\ & \sum_s b(s) = 1 \\ & b(s) \geq 0 & \forall s \\ & b'(s) \geq 0 & \forall s \end{aligned}$$

When a feasible positive solution exists, $\alpha_j$ is added to the set $\widehat{Sw}^k(\alpha_i)$, though in fact, it may not properly be a member of $Sw^k(\alpha_i)$. If no positive solution exists, we know $\alpha_j$ is not in $Sw^k(\alpha_i)$ and it is not added to $\widehat{Sw}^k(\alpha_i)$. This superset of the switch set can be used to derive an upper bound on error.

While the number of constraints of the type $b(M) = b'(M)$ is exponential in the size of the largest marginal, we expect that the number of variables in each marginal for a useful projection scheme will be bounded by a small constant. In this way, the number of constraints can be viewed as constant (i.e., independent of state space size).

Though the above LPs (for both linear approximations and projection schemes) look complex, they are in fact very similar in size to the LPs used for dominance testing in Monahan's pruning algorithm and the Witness algorithm, involving $O(|S|)$ variables and $O(|\aleph^k|)$ constraints. The number of LP variables is exponential in the number of state variables; however, the factored representation of $\alpha$-vectors allows LPs to be structured in such a way that the state space need not be enumerated (i.e., the variables representing the state probabilities can be clustered). Precisely the same structuring is suggested in [2] and implemented in [8]. Thus solving an LP to test if the agent can switch from $\alpha_i$ to $\alpha_j$ has the same complexity as a dominance test in the pruning phase of POMDP solving. However, there are $O(|\aleph^k|^2)$ pairs of $\alpha$-vectors to test for plan switching whereas the pruning phase may require as few as $|\aleph^k|$ dominance tests if no vector is pruned. Hence, in the worst case, switch set generation may increase the running time for solving the POMDP by a factor of $O(|\aleph^k|)$ at each stage $k$.

For a $k$-stage, finite-horizon POMDP, we can now bound the error in decision quality due to approximation $S$. Define the bound on the maximum error introduced at each stage $j$,

when $\alpha$ is viewed as optimal, as:[7]

$$B_S^j(\alpha) = \max_b \max_{\alpha' \in \widehat{Sw}^j(\alpha)} b \cdot (\alpha - \alpha')$$

Since error at a belief state is simply the expectation of the error at its component states, $B_S^j(\alpha)$ can be determined by comparing the vectors in $\widehat{Sw}^j(\alpha)$ with $\alpha$ componentwise (with the maximum difference being $B_S^j(\alpha)$). Let $B_S^j = \max_{\alpha \in \aleph^j} B_S^j(\alpha)$ be the greatest error introduced by a single approximation $S$ at stage $j$. Then the total error for $k$ successive approximations is bounded by $U_S^k = \sum_{i=1}^k \gamma^i B_S^i$. For an infinite-horizon POMDP, assume we have been given the infinite-horizon value function $\aleph^*$ (i.e., no stages are involved). Then we only need to compute the switch sets $Sw^*(\alpha)$ for this single $\aleph$-set, and the maximum one-shot switching error $B_S^*$. The upper bound on the loss incurred by applying $S$ indefinitely is simply $U_S^* = B_S^*/(1-\gamma)$. Computing the error $U_S^*$ is roughly equivalent to performing $O(|\aleph^*|)$ dynamic programming backups on $\aleph^*$.

The LP formulation used to construct switch sets is computationally intensive. Other methods can be used however to construct these switch sets. We have, for example, implemented a scheme whereby belief states are treated as vectors in $\Re^{|S|}$, and projection schemes are viewed as displacing these vectors. The displacement vectors (vectors which when added to a belief state $b$ give $S(b)$) induced by a scheme $S$ can be computed easily and can be used to determine the direction in which belief state approximation shifts the true belief state. This is turn can be used to construct overestimates of switch sets. While giving rise to looser error bounds, this method is much more efficient in practice. Our emphasis, however, is on the analysis of error due to approximation, so we do not dwell on this scheme in this paper (see [16] for details).

### 3.2 Alternative Plans

The cumulative error induced by switching plans at current and future stages can be bounded in a tighter way. The idea is to generate the set of *alternative plans* that may be executed as a result of both current and *future* approximations. Suppose that an agent, due to approximation at stage $k$ changes its belief state from $b$ to $S(b)$. This can induce a change in the choice of optimal $\alpha$-vector in $\aleph^k$, say from $\alpha_1$ to $\alpha_2$. However, even though the agent has switched and chosen the first action associated with $\alpha_2$, it has not necessarily committed to implementing the entire *conditional plan* $CP(\alpha_2)$ associated with $\alpha_2$. This is because further approximation at stage $k-1$ may cause it to switch from the *continuation* of $CP(\alpha_2)$.

Suppose for instance that $CP(\alpha_2) = \langle a; \sigma \rangle$, where $\sigma(z) = \alpha_3 \in \aleph^{k-1}$. If $z$ is observed, and the agent updates its (approximate) belief state $S(b)$ accurately to obtain $S(b)'$, then the maximizing vector at the next stage is necessarily $\alpha_3$. But given that $S(b)'$ will be approximated before the maximizing vector is chosen, the agent may adopt some other continuation of the plan if $\alpha_3$ does not maximize value for the (second) approximated belief state $S(S(b)')$. In fact, the agent may implement $CP(\alpha_4)$ at stage $k-1$ for any $\alpha_4 \in Sw^{k-1}(\alpha_3)$. Notice that the value of the plan actually implemented—doing the first action of $\alpha_2$, followed by the first action of $\alpha_4$, and so on—may not be represented by any $\alpha$-vector in $\aleph^k$.

---

[6] These equations can be generalized for POMDPs with non-binary variables, though giving more than one equation per subset.

[7] We use $\widehat{Sw}^j$ instead of $Sw^j$ to emphasize the fact that we use the approximate switch set generated for a projection scheme; however, all definitions apply equally well to exact switch sets if they are available.

We can actually construct the values of such plans, and thus obtain much tighter error bounds, while we perform dynamic programming. We recursively define the set of *alternative sets*, or *Alt*-sets for each vector at each stage.[8] We first define

$$Alt^1(\alpha) = Sw^1(\alpha)$$

That is, if $\alpha$ is optimal at stage 1, then any vector in its switch set can have its plan executed. The *future alternative set* for any $\alpha \in \aleph^k$, where $CP(\alpha) = \langle a, \sigma \rangle$, is:

$$FAlt^k(\alpha) = \{\alpha(\langle a, \sigma' \rangle) : (\forall z) \, \sigma'(z) \in Alt^{k-1}(\sigma(z))\}$$

If $\alpha$ is in fact chosen to be executed at stage $k$, true expected value may in fact be given by *any* vector in $FAlt^k(\alpha)$, this is due to future switching of policies at stages following $k$. Finally, define

$$Alt^k(\alpha) = \cup\{FAlt^k(\alpha') : \alpha' \in Sw^k(\alpha)\}$$

If $\alpha$ is in fact optimal at stage $k$ for a given belief state $b$, but $b$ is approximated currently and at every future stage, then expected value might be reflected by any vector in $Alt^k(\alpha)$. These vectors correspond to every possible course of action that could be adopted because of approximation: if we switch vectors at stage $k$, we could begin to execute (the plan associated with) any $\alpha' \in Sw^k(\alpha)$; and if we begin executing $\alpha'$, we could end up executing (the plan associated with) any $\alpha'' \in FAlt^k(\alpha')$.

Given these *Alt*-sets, the error associated with belief state approximation can be given by the maximum difference in value between any $\alpha$ and one of its *Alt*-vectors. These *FAlt* and *Alt*-sets can be computed by dynamic programming while a POMDP is being solved. The complexity of this algorithm is virtually identical to that of generating $\aleph^k$ from $\aleph^{k-1}$, with the proviso that there are $|\aleph^k|$ *Alt*-sets. However, these sets grow exponentially much like the sets $\aleph^k$ would if left unpruned. However, these sets can be pruned in exactly the same way as $\aleph$-sets, with the exception that since we want to produce a worst-case bound on error, we want to construct a lower surface for the *Alt*-sets rather than an upper surface.

Given any *Alt*-set, we denote by $\widehat{Alt}$ the collection of vectors that are *anti-dominating* in *Alt*. For example, if the collection of vectors in Figure 4, form the set $Alt^k(\alpha)$, then the vectors $\alpha_1$ and $\alpha_4$, making up the lower surface of this set, form $\widehat{Alt}^k(\alpha)$. $\widehat{FAlt}^k(\alpha)$ is defined similarly. The set of antidominated vectors can be pruned in exactly the same way that dominated $\alpha$-vectors are pruned from a value function. The same structuring techniques can be used to prevent explicit state enumeration as well. This pruning can keep the $\widehat{Alt}$-sets very manageable in size. Assuming we have an approximation $\widehat{Alt}^k(\alpha)$ of $Alt^k(\alpha)$ for every $\alpha \in \aleph^k$, we construct $\widehat{Alt}^{k+1}(\alpha)$ as follows: (a) $Sw^{k+1}(\alpha)$ is constructed for each $\alpha \in \aleph^{k+1}$; (b) $\widehat{FAlt}^{k+1}(\alpha)$ is constructed using $\widehat{Alt}^k(\alpha)$, and is then pruned to retain only anti-dominating vectors; and (c) $\widehat{Alt}^{k+1}(\alpha)$ is defined as the union of the $\widehat{FAlt}^{k+1}(\alpha')$ sets for those $\alpha' \in Sw^{k+1}(\alpha)$, and is then pruned.

---
[8]This definition can be more concisely specified, but this format makes the computational implications clear.

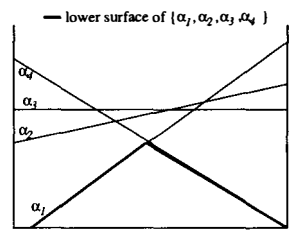

Figure 4: Lower surface

The following quantity bounds the error associated with approximating belief state using scheme $S$ over the course of a $k$-stage POMDP, when $\alpha$ represents optimal expected value for the initial belief state:

$$E_S^k(\alpha) = \max_b \max\{b \cdot (\alpha - \alpha') : \alpha' \in \widehat{Alt}^k(\alpha)\}$$

This error can be computed using simple pointwise comparison of $\alpha$ with each such $\alpha'$. It can also be restricted to that region of belief space where $\alpha$ is optimal; maximizing the difference only over belief states in that area to obtain a tighter bound. Approximation error can be bounded globally using

$$E_S^k = \max\{E_S^k(\alpha) : \alpha \in \aleph^k\}$$

Furthermore, $E_S^k \leq U_S^k$ since alternate vectors provide a much tighter way to measure cumulative error.

For an infinite-horizon problem, we can compute switch sets once as in the computation of $U_S^*$. To compute a tighter bound $E_S^*$, we can construct $k$-stages of $\widehat{Alt}$-sets, backing up from $\aleph^*$. The bound $E_S^k$ is computed as above, and we set

$$E_S^* = E_S^k + \gamma^k U_S^*$$

In this way, we can obtain fairly tight bounds on the error induced by belief state approximation.

## 4 Value-Directed Approximations

The bounds $B^k(\alpha)$ and $E^k$ described above can be used in several ways to determine a good projection scheme. In order to compute error bounds to guide our search for a good projection scheme, our "generic algorithm" will have to determine the error associated with a different projection $S$ applied to each $\alpha$-vector. Because of this, we will consider the use of *different* projection schemes $S_\alpha$ for each $\alpha$-vector (at each stage if we have a finite-horizon problem). Despite the fact that we previously derived bounds on error assuming a uniform projection scheme, our algorithms work equally well (i.e., provide legitimate bounds) if different projections are used with each vector. The projection $S_\alpha$ adopted for vector $\alpha$ simply influences its switch set. Since the agent knows which vector it is "implementing" at any point in time, we can record and easily apply the projection scheme $S_\alpha$ for that vector. This allows the agent to tailor its belief state approximation to provide good results for its currently anticipated course of action. This in turn will lead to much better performance than using a uniform scheme.

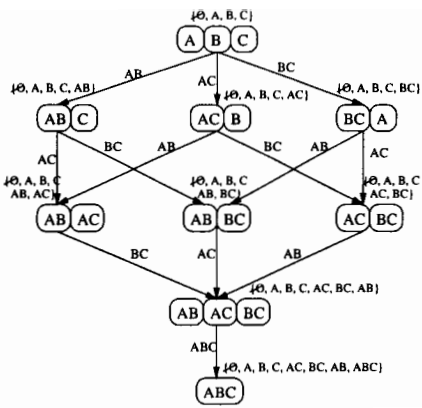

Figure 5: Lattice of Projection Schemes

### 4.1 Lattice of Projection Schemes

We can structure the search for a projection scheme by considering the lattice of projection schemes defined by subset inclusion. Specifically, we say $S_1$ *contains* $S_2$ (written loosely $S_2 \subseteq S_1$) if every subset of $S_2$ is contained within some subset of $S_1$. This means that $S_2$ is a finer "partition" than $S_1$. The lattice of projections for three binary variables is illustrated in Figure 5. Each node represents the set of marginals defining some projection $S$. Above each node, the subsets corresponding to its constraining equations are listed (we refer to each such subset as a *constraint*). The finest projections (which are the "most approximate" since they assume more independence) are at the top of the lattice. Edges are labeled with the subset of variables corresponding to the single constraining equation that must be added to the parent's constraints to obtain the child's constraints.

It should be clear that if $S_2 \subseteq S_1$, then $S_1$ offers (not necessarily strictly) tighter bounds on error when used instead of $S_2$ at any point. To see this, imagine that various approximation schemes are used for different $\alpha$-vectors at different stages, and that $S_2$ is used whenever $\alpha \in \aleph^j$ is chosen. If we keep everything fixed but replace $S_2$ with $S_1$ at $\alpha$, we first observe that $Sw_{S_1}^k(\alpha) \subseteq Sw_{S_2}^k(\alpha)$. This ensures that $B_{S_1}^j(\alpha) \leq B_{S_2}^j(\alpha)$ and $B_{S_1}^j \leq B_{S_2}^j$. If all other projection operators are the same, then obviously $U_{S_1}^k \leq U_{S_2}^k$. Similar remarks apply to the infinite-horizon case. Furthermore, given the definition of *Alt*-sets, reducing the switch set for $\alpha$ at stage $k$ by using $S_1$ instead of $S_2$ ensures that the *Alt*-sets at all preceding stages are no larger (and may well be smaller) than they would be if $S_2$ were used. For this reason, we have that $E_{S_1}^k \leq E_{S_2}^k$ (and similarly $E_{S_1}^* \leq E_{S_2}^*$).

Consequently, as we move down the lattice, the bound on approximation error gets smaller (i.e., our approximations improve, at least in the worst case). Of course, the computational effort of monitoring increases as well. The precise computational effort of monitoring will depend on the structure of the DBN for the POMDP dynamics and its interaction with the marginals given by the chosen projection scheme; however, the complexity of inference (i.e., the dominant factors in the corresponding clique tree), can be easily determined for any node in the lattice.

### 4.2 Search for a Good Projection Scheme

In a POMDP setting, the agent may have a bounded amount of time to make an online decision at each time-step. For this reason, efficient belief-state monitoring is crucial. However, just as solving the POMDP is viewed as an offline operation, so is the search for a good projection scheme. Thus it will generally pay to expend some computational effort to search for a good projection scheme that makes the appropriate tradeoff between decision quality and the complexity of belief state maintenance. For instance, if any scheme $S$ with at most $c$ constraints offers acceptable online performance, then the agent need only search the row of the lattice containing those projection schemes with $c$ constraints. However, the size of this row is factorial in $c$. So instead we use the structure of the lattice to direct our attention toward reasonable projections.

We describe here a generic, greedy, anytime algorithm for finding a suitable projection scheme. We start with the root, and evaluate each of its children. The child that looks most "promising" is chosen as our current projection scheme. Its children are then evaluated, and so on; this continues until an approximation is found that incurs no error (specifically, each switch set is a singleton, as we describe below), or a bound on the size of the projection is reached. We assume for simplicity that at most $c$ constraints will be allowed. The search proceeds to depth $c - n$ in the lattice and at each node, at most $n(c - n)$ children are evaluated, so a total of $O(nc^2 - cn^2)$ nodes are examined. Since $c$ must be greater than $n$—the root node itself has $n$ constraints—we assume $O(nc^2)$ complexity. The structure of the lattice ensures that decision quality (as measured by error bounds) cannot decrease at any step. We note that practical and non-practical projections are included in the lattice. In figure 5, the only non-practical scheme is $S = \{AB, AC, BC\}$. During the search, it doesn't matter if a node corresponding to a non-practical scheme is traversed, as long as the final node is practical. If it is not practical, then the best practical sibling of that node is picked or we backtrack until a practical scheme is found. We also note that since this is a greedy approach, we may not discover the best projection with a fixed number of constraints. However, it is a well-structured search space and other search methods for navigating the lattice could be used.

We first describe one instantiation of this algorithm, the *finite-horizon U-bound search*, for a $k$-stage, finite-horizon POMDP. Given the collections of $\alpha$-vectors $\aleph^1, \cdots, \aleph^k$, we run the following search independently for each vector $\alpha \in \aleph^i$ for each $i \leq k$. The order does not matter; we will end up with a projection scheme $S$ for each $\alpha$-vector, which is applied whenever that $\alpha$-vector is chosen as optimal at stage $i$. We essentially minimize (over $S$) each term $B^i_S(\alpha)$ in the bound $U^k$ independently. For a given vector $\alpha$ at stage $i$, the search proceeds from the root in a greedy fashion. Each child $S$ of the current node is evaluated by computing $B^i_S(\alpha)$, which basically requires that we compute the switch set $Sw^i_S(\alpha)$, which in turn requires the solution of $|\aleph^i|$ LPs. Once the projection schemes $S_\alpha$ for each $\alpha$ are found, the error bound $U^k$ is given by the sum of the bounds $B^i$ as described in the previous section. At each stage $i$, the number of LPs that must be solved is $O(nc^2|\aleph^i|^2)$ since there are $O(|\aleph^i|)$ $\alpha$-vectors and for each $\alpha$-vector, the lattice search traverses $O(nc^2)$ nodes, each requiring the solution of $O(|\aleph^i|)$ LPs. Since the solution of the original POMDP requires the solution of at least $|\aleph|$ LPs, the overhead incurred is at most a factor of $nc^2|\aleph|$.

The method above can be streamlined considerably. When comparing two nodes, it is not always necessary to generate the entire switch set to determine which node has the lowest bound $B^i(\alpha)$. Each vector $\alpha'$ in $\alpha$'s switch set introduces an error of at most $\max_b \{b(\alpha - \alpha')\}$. Since $B^i(\alpha) = \max_{\alpha' \in Sw^i(\alpha)} \{\max_b b(\alpha - \alpha')\}$, we can test vectors $\alpha'$ in decreasing order of contributed error until one vector is found to be in the switch set at one node but not the other. The node that does not include this vector in its switch set has the lowest bound $B^i_S(\alpha)$ (where $S$ is that node's projection scheme). Instead of solving $|\aleph^i|$ pairs of LPs, generally only a few pairs of LPs will be solved.

When testing whether two different schemes $S_1$ and $S_2$ allow switching to some $\alpha$-vector, the LPs to be solved for each scheme are similar, differing only in the constraints dictated by each projection scheme. This similarity can be exploited computationally by using techniques that take advantage of the numerous common constraints if we solve similar LPs "concurrently" (for instance, by solving a stripped down LP that has only the common constraints and using the dual simplex method to account for the extra constraints). Though details are beyond the scope of this paper, these techniques are faster in practice than solving each LP from scratch. The greedy search can take full advantage of these speed-ups: each child has only one additional constraint (compared to its parent), so not only can structure be shared across children, but the parent's solution can be exploited as well. We reiterate that these LPs can also be structured, so state space enumeration is not required. Taken together, these computational tricks don't reduce the worst-case running time of $O(nc^2|\aleph|^2)$ LPs; however in practice it is possible that only $\Omega(nc|\aleph|)$ LPs need be solved, in which case, when integrated with the algorithm to solve the POMDP, the overhead incurred would be a factor proportional to $nc$. A thorough experimentation remains to be done.

There are three variations of the algorithm above. The *infinite-horizon U-bound* algorithm is much like the finite-horizon version. However, we only have one set of $\alpha$-vectors, $\aleph^*$, rather than $k$ sets. Thus we compute far fewer switch sets, and calculate the final bound using the equation for $U^*$. The *finite-horizon E-bound algorithm* is similar to the above algorithm as well. The difference is that we compute *Alt*-sets (or rather approximations to them, $\widehat{Alt}^k_S(\alpha)$) to obtain tighter bounds on error. To do this requires that we compute the projection schemes for the various stages in order, from the last stage back to the first. Once a good scheme has been found for the elements of $\aleph^j$, the $\widehat{FAlt}$-sets can be computed for stage $j+1$ without difficulty (this involves simple DP backups). Then switch sets are computed exactly as above, from which $\widehat{Alt}$-sets, and error bounds, are generated. Finally, the infinite-horizon $E$-bound algorithm proceeds by computing the switch sets for a given projection only once for each vector in $\aleph^*$; but additional DP backups to compute *Alt*-sets (as described in the previous section) are needed to derive tight error bounds.

## 5 Illustrative Example

We describe a very simple POMDP to illustrate the benefits of value-directed approximation, with the aim of demonstrating that minimizing belief state error is not always appropriate when approximate monitoring is used to implement an optimal policy. The process involves only seven stages with only one or two actions per stage (thus at some stages no choice needs to be made), and no observations are involved. Yet even such a simple system shows the benefits of allowing the value function to influence the choice of approximation scheme.

We suppose there is a seven-stage manufacturing process whereby four parts are produced using three machines, $M$, $M1$, and $M2$. Parts *P1*, *P2*, *P3*, and *P4* are each stamped in turn by machine $M$. Once stamped, parts *P1* and *P2* are processed separately (in turn) on machine *M1*, while parts *P3* and *P4* are processed together on *M2*. Machine $M$ may be faulty (*FM*), with prior probability $\Pr(FM)$. When the parts are stamped by $M$, parts *P1* and *P2* may become faulty (*F1*, *F2*), with higher probability of fault if *FM* holds. Parts *P3* and *P4* may also become faulty (*F3*, *F4*), again with higher probability if *FM*; but *F3* and *F4* are both less sensitive to *FM* than *F1* and *F2* (e.g., $\Pr(F1|FM) = \Pr(F2|FM) > \Pr(F3|FM) = \Pr(F4|FM)$). If *P1* or *P2* are processed on machine *M1* when faulty, a cost is incurred; if processed when OK, a gain is had; if not processed (rejected), no cost or gain is had. When *P3* and *P4* are processed (jointly) on $M3$, a greater gain is had if both parts are OK, a lesser gain is had when one part is OK, and a drastic cost is incurred if both parts are faulty (e.g., machine $M3$ is destroyed). The specific problem parameters are given in Table 1.

Figure 6 shows the dependencies between variables for the seven-stage DBN of the example.[9] It is clear with three stages to go, all the variables are correlated. If approximate belief state monitoring is required for execution of the optimal policy (admittedly unlikely for such a simple problem!), a suitable projection scheme could be used.

| Stages to go | Actions | Transitions | Rewards |
|---|---|---|---|
| 7) Stamp P1 | Stamp P1 | only affects F1<br>if FM at previous step<br>then Pr(F1) = 0.8 else Pr(F1) = 0.1 | no reward |
| 6) Stamp P2 | Stamp P2 | only affects F2<br>if FM at previous step<br>then Pr(F2) = 0.8 else Pr(F2) = 0.1 | no reward |
| 5) Stamp P3 | Stamp P3 | only affects F3:<br>if FM at previous step<br>then Pr(F3) = 0.1 else Pr(F3) = 0.05 | no reward |
| 4) Stamp P4 | Stamp P4 | only affects F4:<br>if FM at previous step<br>then Pr(F4) = 0.1 else Pr(F4) = 0.05 | no reward |
| 3) Process/Reject P1 | Process P1 | all variables are persistant | if F1 then 0 else 8 |
|  | Reject P1 | all variables are persistant | 4 for every state |
| 2) Process/Reject P2 | Process P2 | all variables are persistant | if F2 then 0 else 8 |
|  | Reject P2 | all variables are persistant | 4 for every state |
| 1) Process/Reject P3,P4 | Process P3,P4 | all variables are persistant | if F3 & F4 then -2000<br>if ~F3 & ~F4 then 16<br>otherwise 8 |
|  | Reject P3,P4 | all variables are persistant | 3.3 for every state |

Table 1: POMDP specifications for the factory example

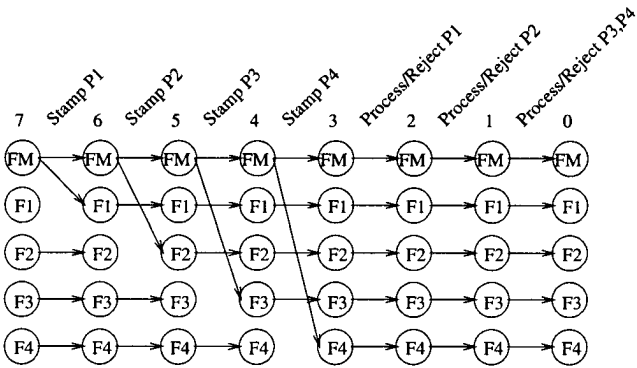

Figure 6: DBN for the factory example

Notice that the decisions to process *P1* and *P2* at stages-to-go 3 and 2 are independent: they depend only on $\Pr(F1)$ and $\Pr(F2)$, respectively, but not on the correlation between the two variables. Thus, though these become quite strongly correlated with five stages to go, this correlation can be ignored without any impact on the decision one would make at those points. Conversely, *F3* and *F4* become much more weakly correlated with three stages to go; but the optimal decision at the final stage *does* depend on their joint probability. Were we to ignore this weak correlation, we run the risk of acting suboptimally.

We ran the greedy search algorithm of Section 4.2 and, as expected, it suggested projection schemes that break all correlations except for *FM* and *F3* with four stages to go, and *F3* and *F4* with three, two, and one stage(s) to go. The latter, $\Pr(F3, F4)$, is clearly needed (at least for certain prior probabilities on *FM*) to make the correct decision at the fi-nal stage; and the former, $\Pr(FM, F3)$, is needed to accurately assess $\Pr(F3, F4)$ at the subsequent stage. Thus we maintain an approximate belief state with marginals involving no more than two variables, yet we are assured of acting optimally.

---

[9] We have imposed certain constraints on actions to keep the problem simple; with the addition of several variables, the problem could easily be formulated as a "true" DBN with identical dynamics and action choices at each time slice.

| Correlation | $L_1$ | $L_2$ | $KL$ | Loss |
|---|---|---|---|---|
| F1/F2 | 0.7704 | 0.3092 | 0.4325 | 1.0 |
| F3/F4 | 0.9451 | 0.3442 | 0.5599 | 0.0 |

Table 2: Comparison of different distance measures

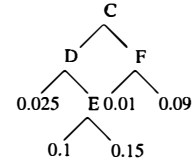

Figure 7: An Example Density Tree

In contrast, if one chooses a projection scheme for this problem by minimizing KL-divergence, $L_1$-distance, or $L_2$-distance, different correlations will generally be preserved. For instance, assuming a uniform prior over *FM* (i.e., machine *M* is faulty with probability 0.5), Table 5 shows the approximation error that is incurred according to each such measure when only the correlation between *F1* and *F2* is maintained or when only the correlation between *F3* and *F4* is maintained. All of these "direct" measures of belief state error prefer the former. However, the loss in expected value due to the former belief state approximation is 1.0, whereas no loss is incurred using the latter. To test this further, we also compared the approximation preferred using these measures over 1000 (uniformly) randomly-generated prior distributions. If only the *F1/F2*-correlation is preserved at the first stage, then in 520 instances a non-optimal action is executed with an average loss of 0.6858. This clearly demonstrates the advantage of using a value-directed method to choose good approximation schemes.

## 6 Framework Extensions

The methods described above provide means to analyze value-directed approximations. Though we focused above on projection schemes, approximate monitoring can be effected by other means. Our framework allows for the analysis of error of any linear approximation scheme $S$. In fact, our analysis is better suited to linear approximations: the constraints on the approximate belief state $S(b)$, if linear, allow us to construct exact switch sets $Sw(\alpha)$ rather than approximations, providing still tighter bounds.

One linear approximation scheme involves the use of density trees [13]. A density tree represents a distribution by aggregation: the tree splits on variables, and probabilities labeling the leaves denote the probability of *every* state consistent with the corresponding branch. For instance, the tree in Figure 7 denotes a distribution over four variables in which states $cd\bar{e}f$ and $cd\bar{e}\bar{f}$ both have probability 0.1. A tree that is polynomially-sized in the number of variables offers an exponential reduction in the number of parameters required to represent a distribution. A belief state can be approximated by forcing it to fit within a tree of a bounded size (or satisfying other constraints). This approximation can be reconstructed at each stage, just like projection. It is clear that a density tree approximation is linear. Furthermore, the number of constraints and required variables in the LP for computing a switch set is small.

We also hope to extend this framework to analyze sampling methods [11, 13, 19]. While such schemes are generally analyzed from the point of view of belief-state error, we would like to consider the impact of sampling on decision quality and develop value-directed sampling techniques that minimize this impact.

## 7 Concluding Remarks

The value-directed approximation analysis we have presented takes a rather different view of belief state approximation than that adopted in previous work. Rather than trying to ensure that the approximate belief state is as close as possible to the true belief state, we try to make the approximate belief state induce *decisions* that are as close as possible to optimal, given constraints on (say) the size of the belief state clusters we wish to maintain. Our approach remains tractable by exploiting recent results on factored representations of value functions.

There are a number of directions in which this research must be taken to verify its practicality. We are currently experimenting with the four bounding algorithms described in section 4.2. Ultimately, although these algorithms provide worst-case bounds on the expected error, it is of interest to gain some insight regarding the *average* error incurred in practice. We are also experimenting with other heuristics, such as the the vector-space method mentioned in Section 3.1, that may provide a tradeoff between the quality of the error bounds and the efficiency of their computation. Other directions include the development of online, dynamic choice of projection schemes for use in search-tree approaches to POMDPs (see, e.g., [14]), as well as solving POMDPs in a bounded-optimal way that takes into account the fact that belief state monitoring will be approximate.

**Acknowledgements** Poupart was supported by NSERC and carried our this research while visiting the University of Toronto. Boutilier was supported by NSERC Research Grant OGP0121843 and IRIS Phase 3 Project BAC.